\documentclass[lettersize,journal]{IEEEtran}
\usepackage{amsmath,amsfonts}
\usepackage{algorithmic}
\usepackage{algorithm}
\usepackage{array}
\usepackage[caption=false,font=normalsize,labelfont=sf,textfont=sf]{subfig}
\usepackage{textcomp}
\usepackage{stfloats}
\usepackage{url}
\usepackage{verbatim}
\usepackage{graphicx}
\usepackage{cite}
\usepackage{amssymb}
\usepackage{multirow}
\usepackage{booktabs}
\usepackage{graphicx}
\usepackage{color,xcolor}
\usepackage{hyperref}
\usepackage{threeparttable}
\usepackage{multirow}
\usepackage{amsmath}
\makeatletter
\renewcommand{\maketag@@@}[1]{\hbox{\m@th\normalsize\normalfont#1}}%
\makeatother

\begin{document}

\title{Joint Depth Estimation and Mixture of Rain Removal From a Single Image}

\author{Yongzhen Wang,
        Xuefeng Yan, Yanbiao Niu, Lina Gong, Yanwen Guo, \textit{Member}, \textit{IEEE}, \\and Mingqiang Wei, \textit{Senior Member}, \textit{IEEE}
\thanks{Y. Wang, X. Yan, Y. Niu, and L. Gong are with the School of Computer Science and Technology, Nanjing University of Aeronautics and Astronautics, Nanjing 210016, China (e-mail: wangyz@nuaa.edu.cn, yxf@nuaa.edu.cn, niuyanbiao@126.com, gonglina@nuaa.edu.cn).}

\thanks{Y. Guo is with the National Key Lab for Novel Software Technology, Nanjing University, Nanjing 210000, China (e-mail: ywguo@nju.edu.cn). }

\thanks{M. Wei is with the Shenzhen Research Institute, Nanjing University of Aeronautics and Astronautics, Shenzhen 518038, China (e-mail: mingqiang.wei@gmail.com).}



}

\markboth{Journal of \LaTeX\ Class Files,~Vol.~14, No.~8, August~2021}%
{Shell \MakeLowercase{\textit{et al.}}: A Sample Article Using IEEEtran.cls for IEEE Journals}


\maketitle

\begin{abstract}
 Rainy weather significantly deteriorates the visibility of scene objects, particularly when images are captured through outdoor camera lenses or windshields. Through careful observation of numerous rainy photos, we have found that the images are generally affected by various rainwater artifacts such as raindrops, rain streaks, and rainy haze,  which impact the image quality from both near and far distances, resulting in a complex and intertwined process of image degradation. However, current deraining techniques are limited in their ability to address only one or two types of rainwater, which poses a challenge in removing the mixture of rain (MOR). In this study, we propose an effective image deraining paradigm for Mixture of rain REmoval, called DEMore-Net, which takes full account of the MOR effect. Going beyond the existing deraining wisdom, DEMore-Net is a joint learning paradigm that integrates depth estimation and MOR removal tasks to achieve superior rain removal. The depth information can offer additional meaningful guidance information based on distance, thus better helping DEMore-Net remove different types of rainwater. Moreover, this study explores normalization approaches in image deraining tasks and introduces a new Hybrid Normalization Block (HNB) to enhance the deraining performance of DEMore-Net. Extensive experiments conducted on synthetic datasets and real-world MOR photos fully validate the superiority of the proposed DEMore-Net. Code is available at  \textcolor{magenta}{ \href{https://github.com/yz-wang/DEMore-Net}{https://github.com/yz-wang/DEMore-Net}}.
\end{abstract}


\section{Introduction}
\IEEEPARstart{R}{ain}, being one of the most ubiquitous weather phenomena, unavoidably induces conspicuous image quality degradation and hinders the performance of various outdoor vision systems \cite{cobreces2009grid,buch2011review,ali2011multiple}. The visibility of the captured scenes is significantly deteriorated by the presence of rain streaks in the atmosphere, as well as the detrimental effect of raindrops that fall onto the camera lens. Moreover, the accumulation of minuscule raindrops in the distant background also engenders a haze-like effect, further exacerbating the visual degradation. This intricate and tangled process of image deterioration, resulting from these three forms of rainwater, is a captivating phenomenon and we designate it as a Mixture of Rain (MOR).

Image deraining aims at restoring the sharp images from their rainy counterparts, which is a challenging ill-posed problem. Existing deraining techniques can roughly fall into two categories: rain streak removal and raindrop removal. Early rain streak removal wisdom commonly exploits hand-crafted priors with empirical observations, such as sparse coding  \cite{kang2011automatic}, low-rank representation \cite{chen2013generalized}, and Gaussian Mixture Model (GMM) \cite{li2016rain}. Notwithstanding the commendable outcomes achieved by the prior-based deraining algorithms, their capacity to model and remove rain is restricted. The advent of deep learning techniques in the field of image enhancement \cite{singh2021principal,zhou2022liner} has led to a proliferation of learning-based approaches for rain streak and raindrop removal \cite{yang2017deep,zhang2018density,wei2019semi,qian2018attentive,luo2021waterdrop,yan2022feature}. Despite producing promising results on various datasets, most of these approaches only focus on addressing one type of rainwater, thereby being insufficient for effectively handling the MOR challenge.

\begin{figure*}[htbp] \centering
	\includegraphics[width=0.95\linewidth]{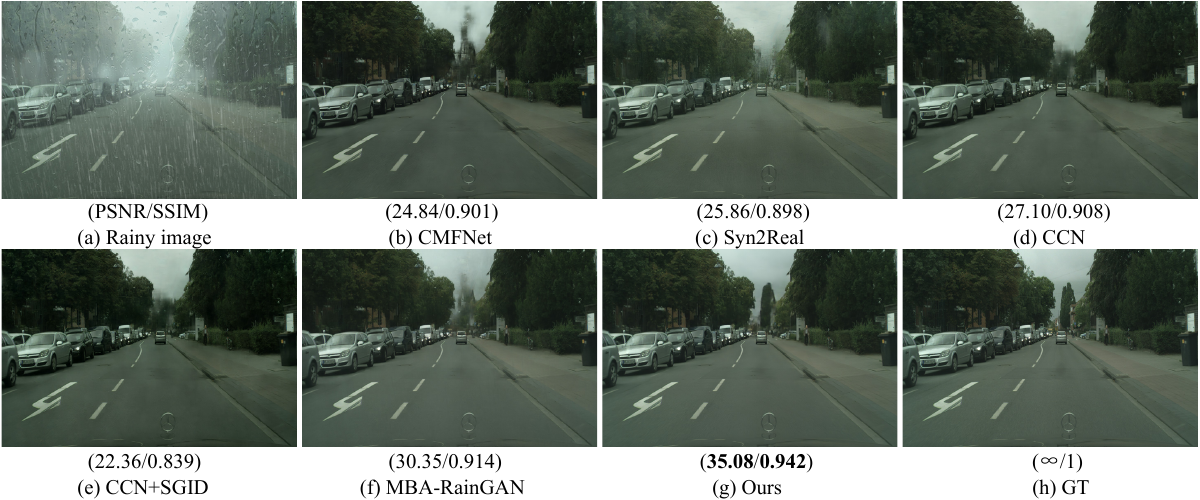}
	\caption{
	Image derained results on a MOR image from the RainCityscapes++ \cite{shen2022mba} dataset. From (a) to (h): (a) the input rainy image, and the deraining results of (b) CMFNet \cite{fan2022compound}, (c) Syn2Real \cite{yasarla2020syn2real}, (d) CCN \cite{quan2021removing}, (e) CCN \cite{quan2021removing} + SGID \cite{bai2022self} (cascading combination method), (f) MBA-RainGAN \cite{shen2022mba}, (g) our DEMore-Net, respectively, and (h) the ground truth.}
	\label{fig:fig1}
\end{figure*}

Recently, this issue has been picked up by several learning-based deraining studies \cite{hu2019depth,guo2020joint,quan2021removing,shen2022mba}. These algorithms argue that more than one type of rainwater is present in rainy images and attempt to remove them in one go. Hu \textit{et al.} \cite{hu2019depth} first observed that rainy images are composed of rain streaks and rainy haze, which often co-occur during image capture. To address this issue, they develop  DAF-Net to remove rain streaks and haze simultaneously. Afterward, several deraining paradigms have been proposed for removing both forms of rainwater artifacts with some success. However, since these efforts only consider two types of rainwater and ignore the MOR effect, they cannot well address the challenging MOR problems. To our knowledge, there is presently only one deraining model that considers the removal of all three forms of rainwater, namely MBA-RainGAN \cite{shen2022mba}. Despite achieving satisfactory results in some rainy scenarios, the deraining capacity of MBA-RainGAN is still limited since it ignores the fact that depth information is quite important, which can provide additional beneficial information to guide the network to remove different forms of rainwater. As illustrated in Fig. \ref{fig:fig1}, compared with the state-of-the-art (SOTA) deraining algorithms, the proposed DEMore-Net with the guidance information of the scene depth produces a much clearer and perceptually more pleasing derained result.


This work goes beyond previous image deraining techniques by considering the importance of depth information and developing a unified deraining paradigm called DEMore-Net that integrates depth estimation and image deraining. DEMore-Net leverages a joint learning framework to simultaneously perform image deraining and depth estimation tasks and encourages them to collaborate and promote each other. The estimated depth map serves as a guide to removing different forms of rainwater in a targeted manner depending on the distance, while the image deraining module helps the depth estimation network learn better depth estimation. Additionally, a novel Hybrid Normalization Block (HNB) is proposed to advance the stability of the network training and enhance its learning and generalization capacities, thereby helping the model in tackling the intractable MOR problem. Moreover, to further advance the MOR removal capacity of the model, an efficient feature enhancement network (self-calibrated convolutions \cite{liu2020improving}) is introduced in the design of our DEMore-Net to generate more discriminative feature representations. Extensive experiments on synthetic MOR testbeds (RainCityscapes++ \cite{shen2022mba}) and real-world MOR photos demonstrate that our DEMore-Net outperforms the SOTA image deraining algorithms significantly. 

Overall, the main contributions are summarized as follows:

\begin{itemize}
	\item A novel unified deraining paradigm is proposed for the removal of mixture of rain, called DEMore-Net, which combines depth estimation and image deraining tasks by a joint learning framework. DEMore-Net is trained in an end-to-end fashion to simultaneously learn about depth estimation and MOR removal, thus encouraging the two sub-tasks to benefit from each other. 
	\item We propose a novel Hybrid Normalization Block (HNB) to enhance the stability of the network during the training phase and facilitate its learning and generalization capacities for better recovery of MOR images.
	\item We compare the proposed DEMore-Net with 14 state-of-the-art image deraining approaches through considerable experiments. The results are evaluated in terms of full- and no-referenced image quality assessments, visual quality, and human subjective surveys. As observed, the proposed DEMore-Net achieves SOTA performance on both synthetic MOR testbeds and real MOR photos.
\end{itemize}

The subsequent sections of this paper are structured as follows. In Section II, a brief review of the relevant literature is provided, which is divided into two categories: single-type rain removal and multi-type rain removal approaches. Section III presents a detailed overview of DEMore-Net, which is proposed for removing the mixture of rain. Section IV describes the implemented experiments and discusses the results, followed by conclusions in Section V.

\section{Related Work}
In this section, we briefly categorize the discussion into two aspects: single-type rain removal and multi-type rain removal algorithms.

\subsection{Single-Type Rain Removal}
\textbf{Rain streak removal.} Conventional rain streak removal efforts resort to exploiting hand-crafted priors based on image statistics to restore the rainy images \cite{kang2011automatic,chen2013generalized,li2016rain,chang2017transformed,zhu2017joint}. Kang \textit{et al.} \cite{kang2011automatic} employ dictionary learning and sparse coding to decompose the rainy images into different components and then reconstruct the clean images. Chen \textit{et al.} \cite{chen2013generalized} develop a generalized low-rank appearance model to remove rain streaks from images. Similarly, Chang \textit{et al.} \cite{chang2017transformed} exploit a low-rank image decomposition model to restore the rain-free images from their rainy versions.  Zhu \textit{et al.} \cite{zhu2017joint} propose a joint optimization framework with three image priors for rain streak removal from rainy images.

Recently, learning-based approaches have demonstrated their superiority for rain streak removal \cite{yang2017deep,zhang2018density,li2018recurrent,zhang2019image,DBLP:journals/tcsv/AhnJK22}. Yang \textit{et al.} \cite{yang2017deep} develop a multi-task framework to detect and remove rain streaks jointly. Zhang \textit{et al.} \cite{zhang2018density} create a density-aware image deraining network to simultaneously predict rain density and remove rain streaks. Li \textit{et al.} \cite{li2018recurrent} develop a recurrent squeeze-and-excitation contextual dilated network for single image deraining and achieve very promising results. Zhang \textit{et al.} \cite{zhang2019image} exploit a conditional GAN-based model with additional regularization for rain streaks removal. Ahn \textit{et al.} \cite{DBLP:journals/tcsv/AhnJK22} develop a two-step rain removal method that first estimates the rain density and rain streak intensity and then remove them.

\textbf{Raindrop removal.} Since most rain streak removal algorithms cannot be directly applied to raindrop removal, many approaches have been proposed for raindrop detection and removal \cite{kurihata2005rainy,roser2009video,qian2018attentive,quan2019deep,luo2021waterdrop,yan2022feature}. Kurihata \textit{et al.} \cite{kurihata2005rainy} employ the well-known PCA to learn the shape of raindrops and attempt to match the rainy regions. Roser \textit{et al.} \cite{roser2009video} propose an algorithm for monocular raindrop detection in single images. Eigen \textit{et al.} \cite{eigen2013restoring} develop a convolutional network to remove raindrops from a single image. Latter, You \textit{et al.} \cite{you2016adherent} exploit Spatio-temporal information for video raindrop removal. More recently, Qian \textit{et al.} \cite{qian2018attentive} propose an attentive GAN-based network for single-image raindrop removal. Quan \textit{et al.} \cite{quan2019deep} create a CNN-based network to remove raindrops by using shape-driven attention and channel re-calibration. Yan \textit{et al.} \cite{yan2022feature} propose a two-stage video-based raindrop removal approach that first employs a single image module to produce the initial clean results and then refine them by using temporal constraints.

\subsection{Multi-Type Rain Removal}
Very recently, several image deraining approaches \cite{hu2019depth,guo2020joint,zhang2021dual,quan2021removing,shen2022mba} have noticed the presence of more than one type of rainwater artifact in rainy images and attempted to remove them in one go. Hu \textit{et al.} \cite{hu2019depth} first establish a new dataset with rain streaks and rainy haze, and then develop an end-to-end model to remove them simultaneously. Guo \textit{et al.} \cite{guo2020joint} propose an integrated multi-task framework to handle the joint raindrop and haze removal problem by combining the atmospheric scattering model and deep neural network. Zhang \textit{et al.} \cite{zhang2021dual} create a dual branch neural network for removing both rain streaks and raindrops. Quan \textit{et al.} \cite{quan2021removing} propose an effective cascaded network for removing raindrops and rain streaks simultaneously via using the neural architecture search algorithm. Shen \textit{et al.} \cite{shen2022mba} develop a multi-branch attention GAN-based framework to simultaneously remove rain streaks, raindrops, and rainy haze, which is the only study that considers the MOR effect among existing rain removal algorithms. However, these methods either ignore the MOR effect or do not exploit the depth information of the scene, which can be regarded as important prior knowledge to guide the network in removing different types of rainwater artifacts, thus limiting their rain removal capacity in these complex scenarios.
\section{DEMore-Net}
As the scene depth can accurately portray the different forms of rainwater artifacts according to distance, which can be used as important prior knowledge to guide the network to remove these artifacts in a targeted manner. Unfortunately, prevailing image deraining methods tend to overlook this significant guidance information, thus limiting their deraining performance in various scenarios. To address this limitation, we propose a unified and efficient deraining paradigm that seamlessly integrates the depth estimation and image deraining tasks for MOR removal, termed DEMore-Net. In the following, we first present the imaging model of MOR images, which is degraded by three forms of rainwater. Next, the overview of DEMore-Net is elaborated to reveal how we address the entangled MOR removal problem. Subsequently, we describe the network architecture of the MOR removal and depth estimation networks, showcasing how they operate in our unified deraining paradigm. Finally, the proposed Hybrid Normalization Block (HNB) is introduced for better restoring the MOR images.

\begin{figure*}[!h] \centering
	\includegraphics[width=1.0\linewidth]{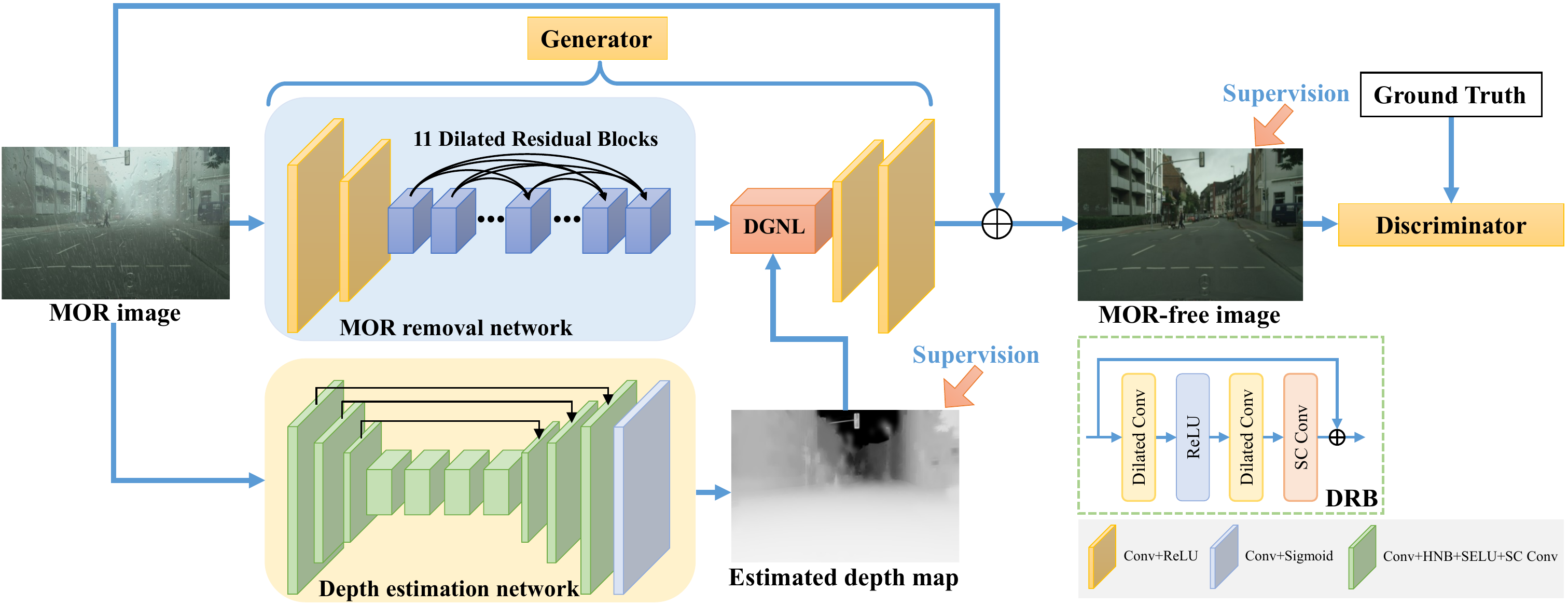}
	\caption{
	Overview of DEMore-Net. It is composed of three components: a generator composed of a MOR removal network and a depth-guided non-local (DGNL) module, a depth estimation network, and a discriminator. DRB refers to dilated residual blocks, HNB refers to Hybrid Normalization Block, and SC Conv denotes self-calibrated convolutions. }
	\label{fig:fig2}
\end{figure*}

\subsection{MOR Imaging Model}
\textbf{Rain Streak Model}. We adopt the widely used rain model \cite{luo2015removing} to define a rain streak degraded image $R_{s}$ as a superposition of the clean background $B$ and the accumulated rain streak layer $S$, and the observed $R_{s}(x)$ at pixel $x$ can be expressed as follows:
\begin{equation}
R_{s}(x)=B(x)+S(x),
\end{equation}
where 
\begin{equation}
S(x)=S_{\text {pattern }}(x) \odot t_{r}(x).
\end{equation}

In the above equations, $S_{\text {pattern }}(x) \in[0,1]$ denotes the intensity of uniformly-distributed rain streaks in the image; $\odot$ refers to the pixel-wise multiplication; and $t_{r}(x)$ denotes the intensity map of rain streak, which depends on the value of the scene depth $d(x)$ and can be expressed as:
\begin{equation}
t_{r}(x)=e^{-\alpha d(x)},
\end{equation}
where $\alpha$ denotes the attenuation coefficient, which controls the intensity of the rain streaks. Specifically, objects close to the camera lens are mainly affected by the raindrops and the rain streak intensity is smaller here. As the objects are far from the camera, rain streak intensity gradually increases to its maximum intensity and then falls to zero with the increase of $d(x)$. At this point, the image is mainly degraded by the rainy haze.

\textbf{Raindrop Model}. A raindrop degraded image $R_{d}$ can be modeled as a combination of the clean background $B$ as well as the impact of the raindrops $D$, and the observed $R_{d}(x)$ can be expressed as \cite{qian2018attentive}:
\begin{equation}
R_{d}(x)=(1-M(x)) \odot B(x)+D(x),
\end{equation}
where $M(x) \in{\{1,0\}}$ denotes whether the pixel $x$ is corrupted by raindrops or not, and $D(x) \in[0,1]$ refers to the raindrop layer. 

\textbf{Rainy Haze Model}. Unlike rain streaks, the visual intensity of rainy haze increases exponentially with the scene depth $d(x)$, we adopt the standard atmosphere scattering model \cite{narasimhan2000chromatic} to simulate the image degradation process caused by rainy haze, and the observed hazy image $R_{h}(x)$ at pixel $x$ can be formulated as:
\begin{equation}
R_{h}(x)={B}(x) t(x)+{A}(1-t(x)),
\end{equation}
where 
\begin{equation}
t(x)=e^{-\beta d(x)}.
\end{equation}

In the above equations, $A$ represents the global atmosphere light, $t(x)$ denotes the atmospheric transmission, and $\beta$ denotes the atmosphere scattering parameter. Note that a larger $\beta$ suggests a thicker rainy haze and vice versa.

\textbf{MOR Model}. In rainy weather, especially under heavy rain conditions, raindrops, rain streaks, and rainy haze often appear simultaneously during outdoor image capture. Therefore, a MOR image $R_{mor}(x)$ can be formulated as follows:
\begin{equation} \label{eq11}
\begin{split}
R_{mor}(x)=&((1-M(x)) \odot(B(x)+S(x))+ \hat{A}D(x))t_x \\
&+ (1-t_x)\hat{A},
\end{split}
\end{equation}
where $\hat{A}$ refers to the light conditions and color cast from active light sources. Unlike the fixed global atmosphere light $A$, $\hat{A}$ enables us to model the nighttime MOR images, which expands the application scenarios of the MOR imaging model. Note that rain streaks and rainy haze may also change the illumination conditions, which in turn affects the transparency of raindrops during image capture. That is, these three rainwater artifacts interact with each other to form the final MOR effect, which is a quite complex and entangled process.

\subsection{Overview of DEMore-Net}
As observed in Eq. \ref{eq11}, scene depth plays a crucial role in the degradation process of MOR images. To fully exploit the informative guidance provided by the scene depth, we present a highly effective and unified deraining paradigm called DEMore-Net. As illustrated in Fig. \ref{fig:fig2}, DEMore-Net consists of two principal modules: the MOR removal sub-network and the depth estimation sub-network. In this way, the joint learning framework enables the MOR removal and depth estimation tasks to collaborate and benefit from each other. DEMore-Net operates in an end-to-end fashion, taking a MOR photo as input and producing the corresponding MOR-free image and depth map as outputs. To produce clearer and more realistic MOR-free images, we introduce the adversarial training strategy in our MOR removal sub-network. 

The pipeline of DEMore-Net can be described in detail as follows: First, the input MOR image undergoes preliminary image restoration and depth estimation tasks through both the MOR removal sub-network and the depth estimation sub-network. Then, the predicted depth map is utilized by the MOR removal sub-network to provide guidance for better restoring the MOR image. To achieve this, a depth-guided non-local (DGNL) \cite{hu2021single} module is employed to fuse the depth information with the coarsely restored features, resulting in a refined MOR-free image. Finally, a discriminator is employed to evaluate the authenticity of the restored MOR-free image, determining whether it is a real clean image or a fake image generated by the MOR removal network. This allows for the generation of more realistic images.

\subsection{Network Architecture}
Overall, DEMore-Net comprises three components: 1) a depth estimation sub-network that employs an encoder-decoder network with multiple feature enhancement modules to produce the depth map; 2) a MOR removal sub-network that produces the MOR-free image using the estimated depth map as additional guidance; and 3) a discriminator that utilizes adversarial training strategies to improve the quality of the restored MOR-free images.

\textbf{Depth Estimation Network}. Inspired by the success of encoder-decoder architecture and ResNet in image restoration field \cite{qian2018attentive,zamir2021multi,wang2022cycle}, we exploit an encoder-decoder network as our depth estimation network and introduce an up-to-date multi-scale feature extraction network (self-calibrated convolutions \cite{liu2020improving}) to further improve the prediction accuracy of depth maps. Self-calibrated convolution module is an improved convolutional network, which can construct long-range spatial and inter-channel dependencies for each spatial location in the feature space, thereby enriching the output features and helping CNNs generate more discriminative representations. Herein, we employ self-calibrated convolutions as our feature enhancement module to improve the depth estimation network’s performance.

As depicted in Fig. \ref{fig:fig2}, the depth estimation network adopts 11 convolutional blocks to extract features from the input MOR image and then output the depth map in an end-to-end manner. For the first 10 convolution blocks, each block contains a convolutional operation, a residual block, a hybrid normalization, a Scaled Exponential Linear Unit (SELU), and a self-calibrated convolution module. Note that all the normalization methods used in this work are the proposed hybrid normalization block, which will be described in the next section. After extracting the image features, we adopt a convolutional operation, and a sigmoid activation function (i.e., $11^{th}$ convolutional block) to output the final depth map in a supervised manner. Here, we adopt the simple $L_1$ loss to train the depth estimation network, which can be expressed as:
\begin{equation}
L_{1}(D)=\frac{1}{W H} \sum_{w=1}^{W} \sum_{h=1}^{H}\left|D(x)^{ w, h}-\hat{y}^{ w, h}\right|_{1},
\end{equation}
where $D(x)$ is the predicted depth map, $\hat{y}$ is the ground truth depth map; $W$ and $H$ denote the width and height of the depth map, respectively.

\textbf{MOR Removal Network}. Considering that adversarial training strategies can encourage the generated images close to the clean images (i.e., ground truth) and make them more realistic, we built the MOR removal sub-network based on a standard generative adversarial network. As demonstrated in Fig. \ref{fig:fig2}, the proposed generator consists of a convolution head to extract image features, a DenseNet-shape body to extract deeper features and restore the clean features simultaneously, and a 1$\times$1 convolution tail to convert channel numbers and output the restored clean image features. The DenseNet-shape body is a stack of dilated residual blocks (DRB) with skip-connections, which contains two 3$\times$3 dilated convolutions and a non-linear ReLU layer. Although applying dilated convolutions can enlarge the fields of view of the model without adding additional computation, it may easily cause the well-known gridding problem. To address this issue, we follow \cite{hu2021single} to combine dilated convolutions with conventional convolutions and set the dilation rates of these 11 DRBs to $\left\{1, 1, 2, 2, 4, 8, 4, 2, 2, 1, 1\right\}$. Finally, the depth-guided non-local (DGNL) module is employed to fuse the predicted clean image features with the depth map, thus helping the model to better remove different types of rainwater according to the distance.

\begin{figure}[htb]
	\centering
	\includegraphics[width=0.75\linewidth]{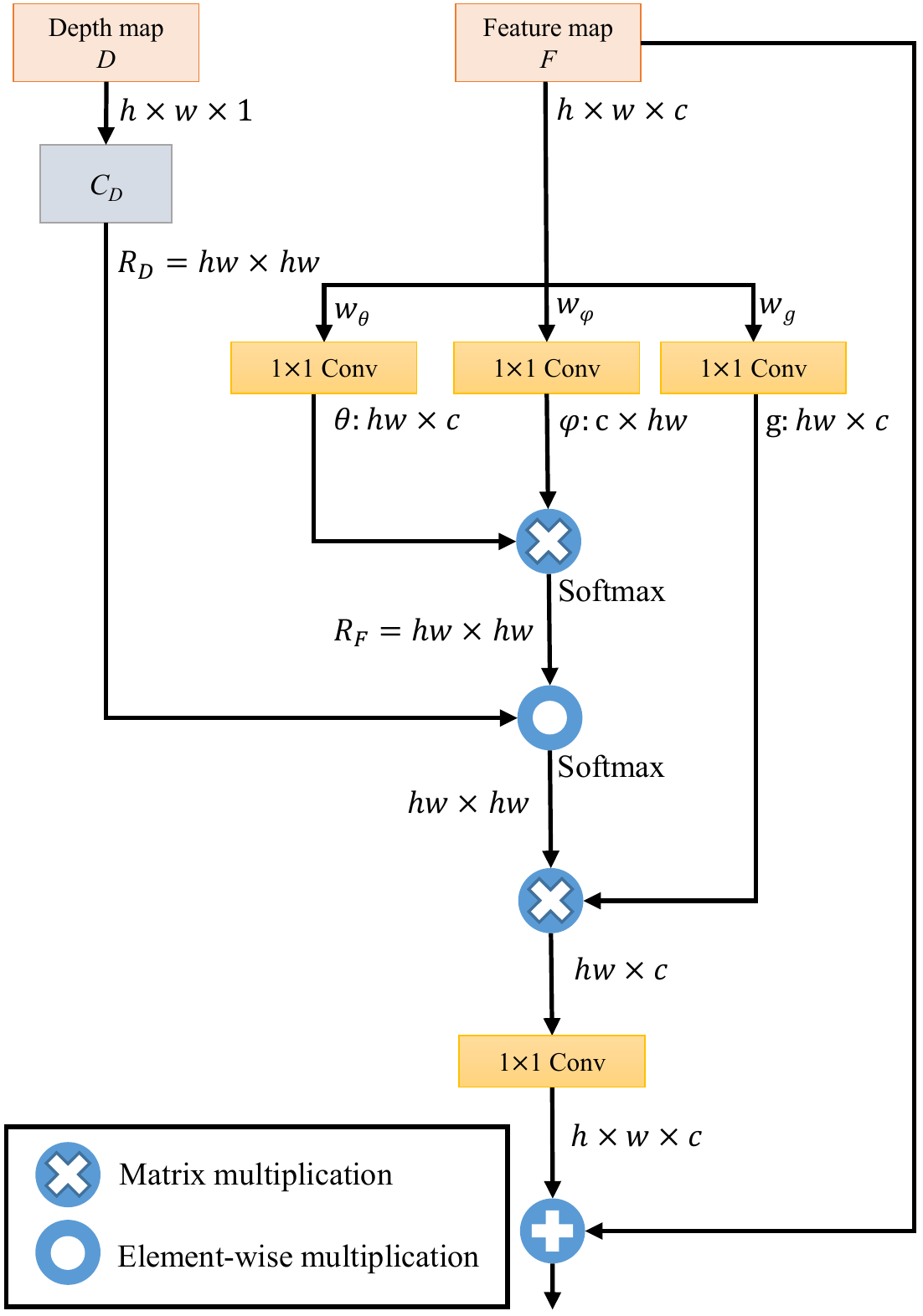}
	\caption{The architecture of the depth-guided non-local network.}
	\label{fig3}
\end{figure}

The function of the discriminator is to determine whether an input image is a real MOR-free image or a fake image produced by the restoration network, thereby encouraging the generator to generate high-quality images. For adversarial training, we adopt 3 convolutional layers and a linear layer to construct our discriminator. Considering that the Least-Squares GAN (LSGAN) loss is more effective in improving the training stability than vanilla GAN loss, we employ the LSGAN to train our MOR removal network. The definition of adversarial loss can be formulated as:
\begin{equation}\label{eq:advg}
\begin{aligned}
L_{adv}(G)=E_{G(x) \sim P_{fake}}[(D(G(x))-1)^2],
\end{aligned}
\end{equation}
\begin{equation}\label{eq:advd}
\begin{split}
L_{adv}(D)=&E_{y \sim P_{real}}[(D(y)-1)^2]\\
&+E_{G(x) \sim P_{fake}}[(D(G(x)))^2],
\end{split}
\end{equation}
where $y$ refers to the ground truth images and $G(x)$ represents the restored MOR-free images. 

\textbf{Depth-Guided Non-Local Module}. The depth-guided non-local (DGNL) module \cite{hu2021single} is an effective feature fusion network, which is employed to build the relationship between each pair of spatial locations based on the depth map and hence enriches the output features. In addition, as mentioned previously, both the type and visual intensity of rainwater artifacts in MOR images depend on scene depth, as does the process of MOR removal. Therefore, we adopt the DGNL module to fuse the depth information with the restored clean image features to help our DEMore-Net better recover the MOR-free images.  

As depicted in Fig. \ref{fig3}, the DGNL module is extended based on the Non-local neural network \cite{wang2018non} and aggregated the depth information as additional guidance to enrich the final output features. In light of this, we regard the depth information as a piece of prior knowledge and leverage the DGNL module to fuse the depth map with restored image features, thus helping DEMore-Net to remove different types of rainwater in a depth-guided manner. After fusing the depth and image features, we employ two convolutional operations to enlarge the feature map and output the final MOR-free images.

\begin{figure}[htb]
	\centering
	\includegraphics[width=0.6\linewidth]{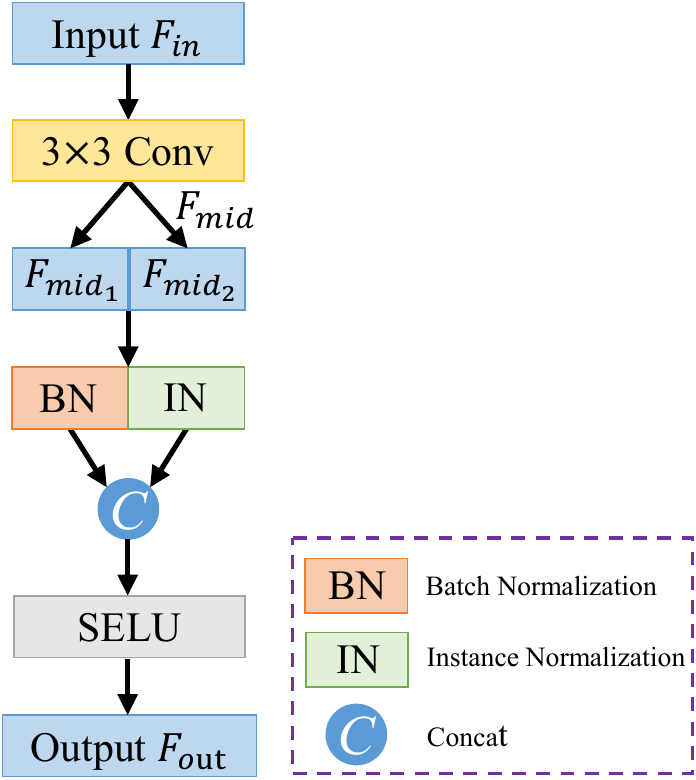}
	\caption{The architecture of the proposed hybrid normalization block (HNB). In our designation, Batch Normalization and Instance Normalization are adopted to preserve content-related features while learning features that are invariant to appearance changes, thus enhancing the MOR removal capacity of the network.}
	\label{fig4}
\end{figure}

\subsection{Hybrid Normalization Block}
Normalization has become a crucial component in various high-level vision tasks but is rarely used in low-level vision tasks, especially Batch Normalization (BN). In fact, the work \cite{pan2018two} has demonstrated that Instance Normalization (IN) can learn features that are not affected by appearance changes, such as brightness, colors, and styles, whilst BN is important to preserve content-related features. Both BN and IN are essential as they can promote each other to benefit the vision tasks. Inspired by this, we carefully integrate BN and IN as building blocks and develop a Hybrid Normalization Block (HNB) to advance the network performance in MOR removal tasks. Specifically, HNB leverages both BN and IN in one normalization operation, thus improving its learning and generalization capacities. Also, the additional parameters and computational costs introduced by HNB can be ignored.

As depicted in Fig. \ref{fig4}, given a feature map $F_{in} \in \mathbb{R}^{C_{in} \times H \times W}$, a 3$\times$3 convolution is first adopted to produce the intermediate feature map $F_{mid} \in \mathbb{R}^{C_{out} \times H \times W}$. Then, HNB divides the input $F_{mid}$ into two parts, namely, $F_{mid_1}$ and $F_{mid_2}$ ($F_{\operatorname{mid}_{1}} \& F_{\text {mid }_{2}} \in \mathbb{R}^{C_{\text {out }} / 2 \times H \times W}$). The first part $F_{mid_1}$ is normalized by BN with learnable parameters and the second part $F_{mid_2}$ is normalized by IN, and then they are concatenated in the channel dimension. HNB employs BN on half of the channels and IN on the other half, so that content-related features can be preserved while learning features that are invariant to appearance changes. After concatenating the two intermediate features, the output feature map $F_{out} \in \mathbb{R}^{C_{out} \times H \times W}$ is obtained by passing features to a SELU layer.

\subsection{Loss Functions}
To train the proposed DEMore-Net more effectively, we take into account all the positive factors that can improve the quality of the restored MOR-free images. Besides depth estimation loss and adversarial loss, we also adopt the reconstruction loss, Dark Channel (DC) loss \cite{li2019semi} and Total Variation (TV) loss \cite{aly2005image} to train our DEMore-Net. The total loss function can be expressed as:
\begin{equation}
\begin{split}
L_{{Total}} = & L_{1}(D) + \lambda_{1} L_{adv}(G)+ L_{rec}+\lambda_{2} L_{dc} \\
&+ \lambda_{3} L_{{tv}},
\end{split}
\end{equation}
where $\lambda_{i}, i=1,2, \cdots 3$, are loss weights, and we set $\lambda_{1}=0.1$, $\lambda_{2}=0.01$, and $\lambda_{3}=0.01$ in our experiments.

\textbf{Reconstruction Loss}. The reconstruction loss in this work is a combination of multi-scale structural similarity (MS-SSIM) loss \cite{wang2003multiscale} and $L_1$ loss, which is formulated as follows:
\begin{equation}
L_{rec}=\alpha \cdot L_{ms{-}ssim} + (1 - \alpha) \cdot L_{1}(R),
\end{equation}
where 
\begin{equation}
\begin{split}
& L_{ms{-}ssim} = 1 - \\ 
& \prod_{m=1}^{M}\left(\frac{2 \mu_{G(x)} \mu_{y}+C_{1}}{\mu_{G(x)}^{2}+\mu_{y}^{2}+C_{1}}\right)^{\beta_{m}}\left(\frac{2 \sigma_{G(x) y}+C_{2}}{\sigma_{G(x)}^{2}+\sigma_{y}^{2}+C_{2}}\right)^{\gamma_{m}}, 
\end{split}
\end{equation}
\begin{equation}
L_{1}(R)=\frac{1}{C W H} \sum_{c=1}^{C} \sum_{w=1}^{W} \sum_{h=1}^{H}\left|G(x)^{c, w, h}-y^{c, w, h}\right|_{1}.
\end{equation}

In the above equations, $\alpha$ is the hyperparameter, and we empirically set $\alpha = 0.1$ in our experiments. $M$ is the total number of the scales and is set to 5 according to \cite{wang2003multiscale}; $\mu_{G(x)}$, $\mu_{y}$ and $\sigma_{G(x)}$, $\sigma_{y}$ represent the mean and standard deviations of $G(x)$ and $y$, $\sigma_{G(x)y}$ refers to their covariance. $\beta_{m}$ and $\gamma_{m}$ denote the relative importance of the two components, both are simply set to 1 in our experiments. Two small constants $C_{1} = 0.0001$ and $C_{2} = 0.0009$ are added to avoid the unstable case of division by zero. 

\textbf{Dark Channel Loss}. In light of the effectiveness of the dark channel prior \cite{he2011single} in the image dehazing field, we apply the Dark Channel loss to help the DEMore-Net deal with the rainy haze in the MOR images, which is formulated as:
\begin{equation}
L_{d c}=\left\|\mathbf{D}_{G \left(x\right)}\right\|_{1},
\end{equation}
where $\mathbf{D}_{G \left(x\right)}$ represents the vector form of the dark channel of the restored MOR-free images $G\left(x\right)$. Since dark channel operation is a highly non-convex and non-linear term, it cannot be directly embedded into the learning networks. We adopt the look-up table scheme (following \cite{li2019semi}) to implement the forward and backward steps of the dark channel operation.

\textbf{Total Variation Loss}. We adopt the total variation loss \cite{aly2005image} to reduce the difference between adjacent pixel values to make the restored images more natural, which can be expressed as:
\begin{equation}
L_{t v}=\left\|\nabla_{h} G(x)+\nabla_{v} G(x)\right\|_{1},
\end{equation}
where $\nabla_{h}$ and $\nabla_{v}$ denote the horizontal and vertical differential operation matrices, respectively.

\section{Experiments}
In this section, comprehensive experiments are conducted on both synthetic RainCityscapes++ benchmark \cite{shen2022mba} and real-world rainy images to evaluate the MOR removal capacity of DEMore-Net and other algorithms. Moreover, we also perform an ablation study to evaluate the effectiveness of each component in DEMore-Net. The details are as follows.

\subsection{Implementation Details}
\textbf{Dataset}. To the best of our knowledge, the RainCityscapes++ dataset \cite{shen2022mba} stands as the sole synthetic MOR benchmark to encompass raindrops, rain streaks, and rainy haze in one consolidated dataset. It comprises a total of 8580 synthetic MOR images, out of which 7580 images serve for training and 1000 images for testing. Henceforth, we adopt the RainCityscapes++ dataset to train and evaluate the performance of our proposed DEMore-Net.

\textbf{Training Details}. DEMore-Net is implemented by PyTorch 1.7 on a system with an Intel Core i9-12900KF CPU, 32 GB RAM, and an NVIDIA GeForce RTX 3090 GPU. We adopt the Adam optimizer with a batch size of 2 to train our model, where the momentum parameters $\beta_1$ and $\beta_2$ are set to 0.9 and 0.999. The total number of iterations and the initial learning rate are empirically set to $300,000$ and $5\times10^{-4}$, respectively. Considering that the original image size of the RainCityscapes++ dataset is too large ($2048\times1024$ pixels), we resize the images to $720\times480$ pixels for both training and inference.

\textbf{Evaluation Settings}. To quantitatively assess the performance of DEMore-Net, we employ the average Peak Signal to Noise Ratio (PSNR) and Structural Similarity index (SSIM) as the evaluation metrics, which are the most widely used image objective evaluation indexes in the field of image restoration. The proposed DEMore-Net is compared with various SOTA image deraining approaches. These algorithms can be divided into two categories, namely, single-type rain removal methods and multi-type rain removal methods. For single-type rain removal methods, we compare with both raindrop, rain streak, and rainy haze removal approaches, including AttentiveGAN \cite{qian2018attentive}, CMFNet \cite{fan2022compound}, MSPFN \cite{jiang2020multi}, Syn2Real \cite{yasarla2020syn2real}, DGCN \cite{fu2021rain}, MPRNet \cite{zamir2021multi}, LD-Net \cite{ullah2021light}, and SGID \cite{bai2022self}. For multi-type rain removal methods, we compare with DAF-Net \cite{hu2019depth}, DGNL-Net \cite{hu2021single}, CCN \cite{quan2021removing}, MBA-RainGAN \cite{shen2022mba}, and the cascade of different types rain removal models (i.e., \textit{raindrop}+\textit{rain streak}+\textit{rainy haze} removal).

\begin{table*}[]
\centering
\caption{Comparison of DEMore-Net with SOTA image deraining algorithms on the RainCityscapes++ test set. RainD, RainS, and RainH are abbreviations of Raindrop, Rain Streak, and Rainy Haze, respectively.} 
\begin{threeparttable}
	\centering
			\begin{tabular}{cccccccc}
				\toprule
				\multirow{1}{*}{Method}&
                \multirow{1}{*}{Publication} &
                \multirow{1}{*}{Removal Type}&
                \multirow{1}{*}{PSNR$\uparrow$} &
                \multirow{1}{*}{SSIM$\uparrow$} & \cr
				\midrule
                AttentiveGAN \cite{qian2018attentive} & CVPR'18 & Raindrop & 24.03 & 0.842    \cr 
                CMFNet \cite{fan2022compound} & arXiv'22 & Raindrop & 28.59 & 0.891  \cr
                MSPFN \cite{jiang2020multi} & CVPR'20 & Rain Streak & 22.19 & 0.704  \cr
                Syn2Real \cite{yasarla2020syn2real} & CVPR'20 & Rain Streak & 27.15 & 0.898  \cr
                DGCN \cite{fu2021rain} & AAAI'21 & Rain Streak & 23.64 & 0.794    \cr
                MPRNet \cite{zamir2021multi} & CVPR'21 & Rain Streak & 22.87 & 0.762   \cr
                LD-Net \cite{ullah2021light} & TIP'21 & Rainy Haze & 20.06 & 0.741    \cr 
                SGID \cite{bai2022self} & TIP'22 & Rainy Haze & 21.87 & 0.761   \cr
                \midrule
                DAF-Net \cite{hu2019depth} & CVPR'19 & RainS + RainH & 25.21 & 0.852   \cr 
                DGNL-Net \cite{hu2021single} & TIP'21 & RainS + RainH & 26.89 & 0.864   \cr 
                CCN \cite{quan2021removing} & CVPR'21 & RainD + RainS & 27.51 & 0.901   \cr    
                MBA-RainGAN \cite{shen2022mba} & ICASSP'22 & RainD + RainS + RainH & 29.16 & 0.913 \cr   
                DGNL-Net + CMFNet & - & RainD + RainS + RainH & 24.89 & 0.866   \cr
                CCN + SGID & - & RainD + RainS + RainH & 24.43 & 0.867   \cr 
                DEMore-Net & ours & RainD + RainS + RainH & \textbf{35.46} & \textbf{0.959}  \cr
                \bottomrule
			\end{tabular}
	\end{threeparttable}
\label{table2}
\end{table*}

\subsection{Comparison with State-of-the-arts}
\begin{figure*}[!h] \centering
	\includegraphics[width=1.0\linewidth]{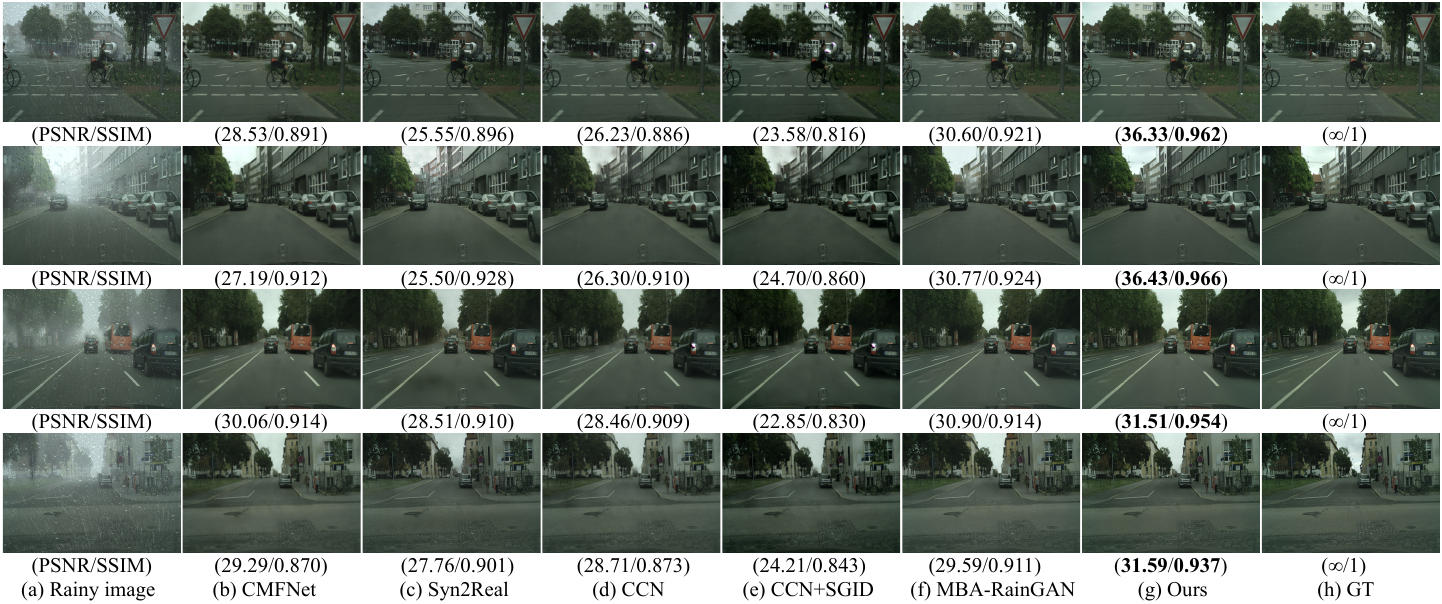}
	\caption{
	Image deraining results on the RainCityscapes++ dataset. From (a) to (h): (a) the rainy image, and the deraining results of (b) CMFNet \cite{fan2022compound}, (c) Syn2Real \cite{yasarla2020syn2real}, (d) CCN \cite{quan2021removing}, (e) CCN + SGID, (f) MBA-RainGAN \cite{shen2022mba}, (g) our DEMore-Net, respectively, and (h) the ground truth. As observed, the proposed DEMore-Net can produce high-quality MOR-free images with well-preserved details.}
	\label{fig5}
\end{figure*}

\textbf{Comparison on Synthetic Dataset}. For quantitative evaluation, we report the averaged PSNR and SSIM metrics of 14 state-of-the-art dehazing algorithms on the RainCityscapes++ test set in Table \ref{table2}. For a fair comparison, all compared methods are retrained on the RainCityscapes++ training set. As can be seen, since single-type rain removal algorithms can only remove one type of rainwater artifact, they cannot achieve satisfactory results. Surprisingly, cascading combinations of different rain removal methods (\textit{raindrop}+\textit{rain streak}+\textit{rainy haze}) also usually yielded poor results. We argue this is because different restoration methods may introduce additional artifacts or lose details in the image processing process (see Fig. \ref{fig5}), and this process is constantly accumulating, degrading the quality of the final restored images. Compared with these SOTA approaches, the proposed DEMore-Net achieves the best performance with 35.46 PSNR and 0.959 SSIM.

\begin{figure*}[!h] \centering
	\includegraphics[width=1.0\linewidth]{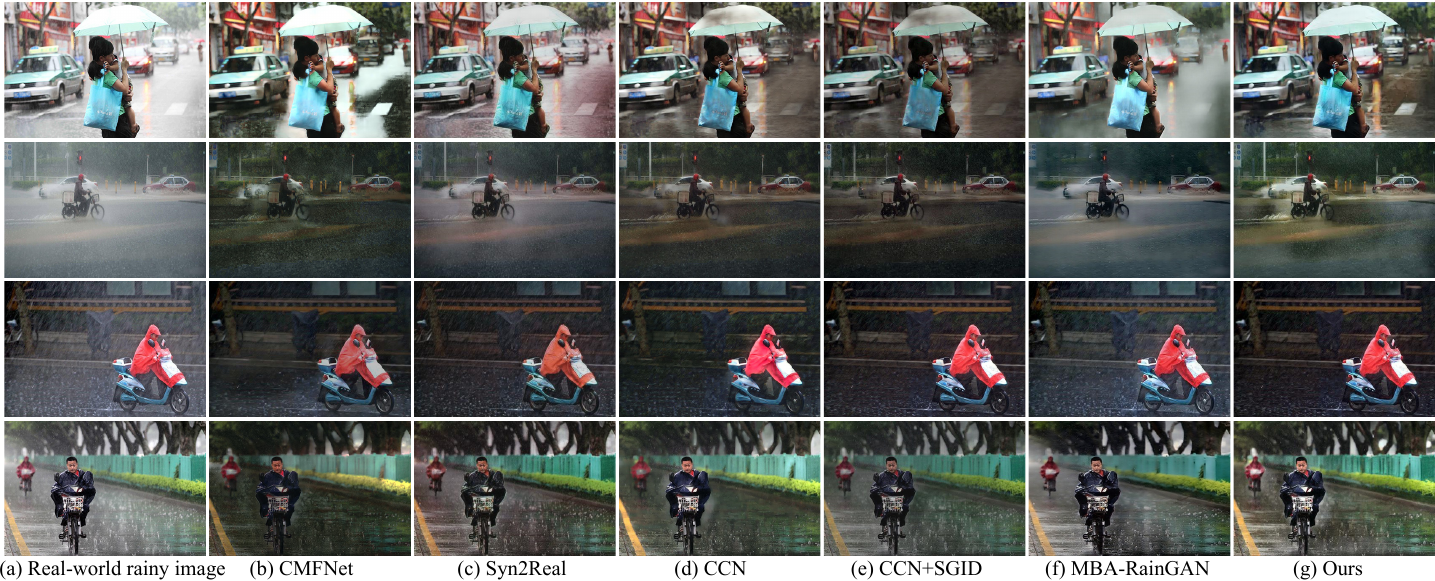}
	\caption{
	Image deraining results on the real-world rainy images. Our DEMore-Net can generate both MOR-free and more natural images.}
	\label{fig6}
\end{figure*}

Moreover, Fig. \ref{fig5} exhibits a visual comparison of synthetic MOR images from the RainCityscapes++ dataset. As observed, Syn2Real cannot completely remove the rainy haze in the image, and some haze remains in the sky area. Although CMFNet and CCN succeed in removing the rain to some extent, they cannot preserve the details of the image well, making the image look a little blurry. CCN+SGID (cascading combination method) seems unable to cope with the intractable MOR problem by combining it with another dehazing algorithm. Compared with the original CCN approach, it introduces additional artifacts and tends to darken the restored images. The deraining results of MBA-RainGAN are pretty good but still cannot completely remove the rainy haze in sky regions. In contrast, our DEMore-Net can produce much clearer and more natural MOR-free images.

\textbf{Comparison on Real-World MOR Images}. To evaluate the effectiveness of DEMore-Net in real-world rainy scenarios, we compare our method with SOTA rain removal approaches on real-world MOR images. Fig. \ref{fig6} displays 4 real-world rainy images and the deraining results by different algorithms. As observed, Syn2Real cannot produce satisfactory deraining results due to color distortion. For CMFNet and CCN, although they have largely overcome the issue of color distortion, they introduce additional artifacts into the image, making the deraining results look unreal and unnatural. The results of CCN+SGID look very dark, and there is still some remaining rain. MBA-RainGAN can remove most of the rain streaks and raindrops in the image, but still cannot effectively address the rainy haze in some regions. Compared with these SOTA methods, our DEMore-Net produces the most realistic MOR-free with perceptually pleasing and high quality.

To better understand the deraining capacity of DEMore-Net in real-world rainy scenarios, a user study is performed to quantitatively evaluate the deraining results. To be more specific, we first prepare 50 real-world rainy images from an existing real-world rainy dataset (DDN-SIRR \cite{wei2019semi}) or from the Internet. Then, we apply 5 representative SOTA deraining algorithms and our DEMore-Net to restore these 50 images. Next, 10 participants (5 males/5 females) are recruited and asked to score the deraining results on a scale from 1 (worst) to 5 (best). We exhibit to each participant these 300 derained images in a random order, without telling them the corresponding deraining algorithms. The experimental results are depicted in Table \ref{Tab666}, demonstrating that our DEMore-Net can cope well with real-world MOR removal tasks.

\begin{table}[htbp]
	\centering
	\caption{User study results. Mean ratings are given ± standard deviation for each deraining algorithm.}
	\label{Tab666}
	\begin{threeparttable}
		\centering
		\setlength{\tabcolsep}{0.8mm}{
			\begin{tabular}{cc}
				\toprule
				\multirow{1}{*}{Method}&
                \multirow{1}{*}{Rating (mean \& standard dev.)} \cr
				\midrule
				CMFNet \cite{fan2022compound} & 3.34 ± 0.45  \cr
				Syn2Real \cite{yasarla2020syn2real}  & 3.16 ± 0.55  \cr
				CCN \cite{quan2021removing} & 3.13 ± 0.74  \cr
				CCN + SGID & 2.75 ± 0.82 \cr
				MBA-RainGAN \cite{shen2022mba} & 3.42 ± 0.57 \cr
				DEMore-Net (ours)   & \textbf{3.74} ± \textbf{0.33}  \cr
				\bottomrule
			\end{tabular}
		}
	\end{threeparttable}
\end{table}

\begin{table}[htbp]
	\centering
	\caption{Quantitative evaluations (NIQE/SSEQ/BRISQUE/PI) with SOTA deraining approaches on 50 real-world rainy images. \textcolor{red}{Red} and \textcolor{blue}{blue} indicate the $1^{st}$ and $2^{nd}$ ranks, respectively.}
	\label{777}
	\begin{threeparttable}
		\centering
		\setlength{\tabcolsep}{0.9mm}{
			\begin{tabular}{ccccc}
				\toprule
				\multirow{1}{*}{Method}&
				\multirow{1}{*}{NIQE$\downarrow$}&
				\multirow{1}{*}{SSEQ$\downarrow$}&
			    \multirow{1}{*}{BRISQUE$\downarrow$}&
				\multirow{1}{*}{PI$\downarrow$}\cr
				\midrule
				Rainy image  & 4.582 & 28.456 & 23.276 & 3.259 \cr
				Syn2Real \cite{yasarla2020syn2real}  & \textcolor{blue}{3.894} & 28.559 & 19.142 & \textcolor{blue}{2.594} \cr
				CMFNet \cite{fan2022compound}  & 4.495 & 27.322 & 21.934 & 3.144 \cr
				CCN \cite{quan2021removing}  & 3.961 & 26.382 & 22.574 & 3.106 \cr
				CCN + SGID  & 4.553 & 28.450 & 23.052 & 3.077 \cr
				MBA-RainGAN \cite{shen2022mba}  & 4.409 & \textcolor{blue}{19.626} & \textcolor{red}{17.938} & 2.939 \cr
				DEMore-Net & \textcolor{red}{3.714} & \textcolor{red}{19.480} & \textcolor{blue}{18.875} & \textcolor{red}{2.555} \cr
				\bottomrule
			\end{tabular}
		}
	\end{threeparttable}
\end{table}

For quantitative comparison, 4 well-known no-reference image quality assessment indexes are employed to evaluate the performance of real-world MOR removal, including NIQE, SSEQ, BRISQUE, and PI \cite{DBLP:conf/eccv/BlauMTMZ18}. All these indexes are evaluated on the above-mentioned 50 real-world rainy images. As tabulated in Table \ref{777}, our DEMore-Net achieves the best performance in terms of NIQE, SSEQ, and PI, showing that the restored MOR-free images produced by our approach are much clearer and more realistic. Furthermore, DEMore-Net also achieves an impressive performance in BRISQUE. In a nutshell, our DEMore-Net wins three of the four indexes, fully validating the superiority of DEMore-Net on real-world rain removal tasks.

\subsection{Ablation Study} \label{ablation}
The proposed DEMore-Net exhibits superior MOR removal capacity compared to 14 SOTA deraining algorithms. To evaluate the effectiveness of DEMore-Net, an ablation study is performed to validate different components, including the DenseNet structure, self-calibrated convolutions, depth estimation branch, adversarial training strategy, and Hybrid Normalization Block (HNB).

We first adopt 11 dilated residual blocks (DRB) with 2 convolutional operations to construct the base model of the MOR removal network (without depth estimation branch) and then train this model via the aforementioned implementation details. The normalization approach used here is group normalization \cite{wu2018group}. Next, different components are gradually added to our base model:
\begin{enumerate}
\item  base model + DenseNet structure $\rightarrow$ $V_1$, 
\item  $V_1$ +  self-calibrated convolutions $\rightarrow$ $V_2$,
\item  $V_2$ +  depth estimation branch (including DGNL) $\rightarrow$ $V_3$,
\item  $V_3$ + adversarial training strategy $\rightarrow$ $V_4$,
\item  $V_4$ + Hybrid Normalization Block $\rightarrow$ $V_5$ (full model).
\end{enumerate}
All these models are retrained in the same training strategy as before and evaluated on the RainCityscapes++ dataset. The experimental results of these models are depicted in Table \ref{table3}.  

As exhibited in Table \ref{table3}, each component in our DEMore-Net contributes to MOR removal, especially the well-designed depth estimation branch, which achieves a 4.96dB PSNR and 0.057 SSIM improvement over variant $V_2$. Additionally, the proposed HNB has significantly advanced the performance of our network in terms of PSNR, and the use of adversarial training strategy and self-calibrated convolutions improve the model performance in SSIM. The adoption of the DenseNet structure has also improved the deraining performance of the model. As observed, the committee with these five components produced the best MOR removal performance, indicating that the five components can complement each other.

\begin{table}[]
\centering
\caption{Ablation study on DEMore-Net. As observed, $V_5$ (our full model) achieves the best performance.} 
\begin{threeparttable}
		\centering
		\setlength{\tabcolsep}{1.3mm}{
			\begin{tabular}{ccccccc}
				\toprule
				\multirow{1}{*}{Variants}&
				\multirow{1}{*}{Base}&
				\multirow{1}{*}{$V_1$}&
				\multirow{1}{*}{$V_2$}&
				\multirow{1}{*}{$V_3$}&
				\multirow{1}{*}{$V_4$}&
				\multirow{1}{*}{$V_5$} \cr
				\midrule
                DenseNet & w/o & \checkmark & \checkmark & \checkmark & \checkmark & \checkmark \cr 
                SC Conv & w/o & w/o & \checkmark & \checkmark & \checkmark & \checkmark \cr 
                Depth Estimation Branch & w/o & w/o & w/o & \checkmark & \checkmark & \checkmark \cr
                Adversarial Training & w/o & w/o & w/o & w/o & \checkmark & \checkmark \cr
                HNB & w/o & w/o & w/o & w/o & w/o & \checkmark \cr
                \midrule
                PSNR & 24.38 & 25.60 & 26.86 & 31.82 & 33.43 & \textbf{35.46} \cr
                SSIM & 0.845 & 0.855 & 0.869 & 0.926 & 0.941 & \textbf{0.959} \cr
                \bottomrule
			\end{tabular}
		}
\end{threeparttable}
\label{table3}
\end{table}

\subsection{Application}
\begin{figure*}[!h] \centering
	\includegraphics[width=1.0\linewidth]{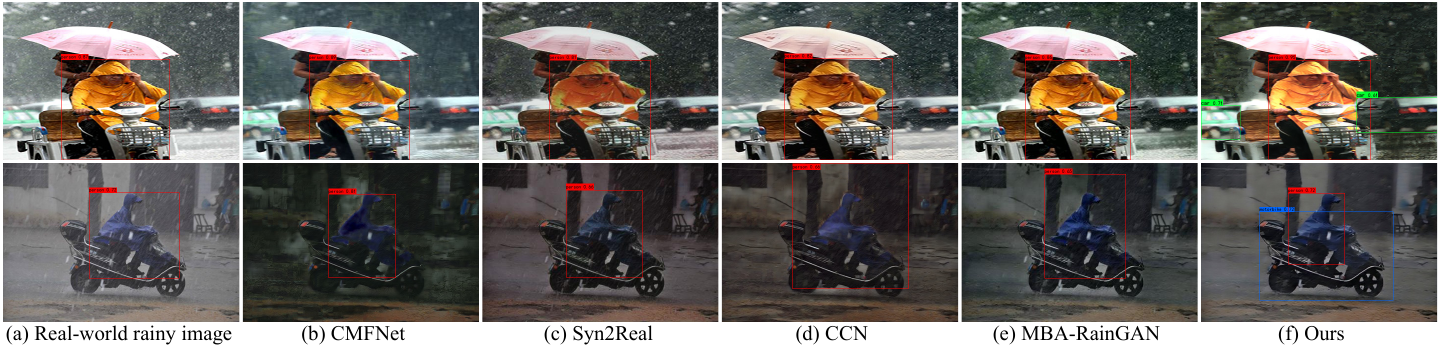}
	\caption{
Object detection results on the real-world rainy images and images restored by different deraining algorithms.}
	\label{fig:fig8}
\end{figure*}

Typically, the accuracy of object detectors drops significantly in rainy conditions due to obvious visibility degradation. To prove that our DEMore-Net can benefit vision-based applications, we adopt a pre-trained YOLOXs \cite{ge2021yolox} detector to detect objects on real-world rainy images and the corresponding deraining results by different algorithms. As depicted in Fig. \ref{fig:fig8}, after removing the rain, the confidence in recognizing objects is greatly improved. Clearly, DEMore-Net outperforms the other image deraining approaches, fully demonstrating the effectiveness of our approach in real-world deraining tasks.

\subsection{Efficiency Analysis}
Considering that efficiency is quite crucial for computer vision systems, we evaluate the computational performance of some typical deraining algorithms and report their average running times in Table \ref{tab4}. All the methods are implemented on an NVIDIA GeForce RTX 3090 GPU. It can be observed that DEMore-Net takes on average about 0.07$s$ to process a rainy image from the RainCityscapes++ dataset (720$\times$480 pixels). Our DEMore-Net has the second-fastest inference speed among the 13 representative deraining approaches.

\begin{table}[htbp]
\centering
	\caption{Average runtime of different deraining algorithms tested on the RainCityscapes++ dataset.}
	\label{tab4}
	\begin{threeparttable}
	\centering
			\begin{tabular}{ccc}
				\toprule
				\multirow{1}{*}{Method}&
                \multirow{1}{*}{Platform} &
                \multirow{1}{*}{Average time} \cr
				\midrule
				AttentiveGAN \cite{qian2018attentive}   & PyTorch (GPU) & 0.25$s$  \cr
				CMFNet \cite{fan2022compound} & PyTorch (GPU) & 0.22$s$  \cr
				MSPFN \cite{jiang2020multi}  & TensorFlow (GPU) & 0.68$s$ \cr
				Syn2Real \cite{yasarla2020syn2real} & PyTorch (GPU) & 0.37$s$ \cr
				DGCN \cite{fu2021rain} & TensorFlow (GPU) & 0.28$s$ \cr
				MPRNet \cite{zamir2021multi} & PyTorch (GPU) & 0.23$s$ \cr  
				LD-Net \cite{ullah2021light}  & PyTorch (GPU) & 0.35$s$  \cr
				SGID \cite{bai2022self} & PyTorch (GPU) & 0.52$s$  \cr
				DAF-Net \cite{hu2019depth} & Caffe (GPU) & \textcolor{red}{0.05$s$}  \cr
				DGNL-Net \cite{hu2021single} & PyTorch (GPU) & 0.09$s$  \cr
				CCN \cite{quan2021removing} & PyTorch (GPU) & 0.36$s$  \cr
				MBA-RainGAN \cite{shen2022mba} & PyTorch (GPU) & 0.43$s$  \cr
				DEMore-Net (ours)   & PyTorch (GPU) & \textcolor{blue}{0.07$s$}  \cr
				\bottomrule
			\end{tabular}
	\end{threeparttable}
\end{table}

\section{Conclusion}
This paper presents a highly efficient unified paradigm for MOR removal, named DEMore-Net. The proposed methodology leverages a joint learning framework to perform depth estimation and image deraining tasks simultaneously. By considering the depth map as an essential prior knowledge, the network is guided to remove various types of rainwater more effectively. Additionally, the deraining module's output features are shared to improve the depth prediction in the other branch, resulting in a collaborative and mutually beneficial approach. To enhance the model's stability and generalization abilities, a novel Hybrid Normalization Block (HNB) is developed in the network design. Moreover, to further improve the MOR removal capacity of DEMore-Net, the self-calibrated convolution network is introduced as a feature enhancement module to produce more discriminative feature representations. Both quantitative and qualitative evaluations demonstrate that DEMore-Net surpasses 14 contemporary SOTA image deraining algorithms.

\bibliographystyle{IEEEtran}
\bibliography{reference}

\vfill

\end{document}